\documentclass[11pt,a4paper]{article}
\usepackage{times,latexsym}
\usepackage{url}
\usepackage[T1]{fontenc}

\usepackage[acceptedWithA]{tacl2021v1}

\usepackage{xspace,mfirstuc,tabulary}

\newcommand{\ex}[1]{{\sf #1}}

\newif\iftaclinstructions
\taclinstructionsfalse 
\iftaclinstructions
\renewcommand{\confidential}{}
\renewcommand{\anonsubtext}{(No author info supplied here, for consistency with
TACL-submission anonymization requirements)}
\newcommand{\instr}
\fi

\iftaclpubformat 

\else

\fi

\usepackage{comment}
\usepackage{multirow}
\usepackage{tikz}
\usepackage{linguex}
\usepackage{booktabs}
\usepackage{xcolor}
\newcommand{\ra}[1]{\renewcommand{\arraystretch}{#1}}

\title{Abstractive Meeting Summarization: A Survey}

\author{
Virgile Rennard$^{1,2}$\Thanks{Primary, equal contribution; $^\dagger$significant contributions.} , Guokan Shang$^{1*}$, Julie Hunter$^{1\dagger}$, Michalis Vazirgiannis$^2$
  \\
  \ \\
  $^1$Linagora, France $^2$\'Ecole Polytechnique, France
  \\
  \texttt{virgile@rennard.org guokan.shang@polytechnique.edu}
  \\
  \texttt{jhunter@linagora.com mvazirg@lix.polytechnique.fr}
}

\date{}

\begin{document}
\maketitle
\begin{abstract}

    A system that could reliably identify and sum up the most important points of a conversation   would be valuable in a wide variety of real-world contexts, from business meetings to medical consultations to customer service calls. Recent advances in deep learning, and especially the invention of encoder-decoder architectures, has significantly improved language generation systems, opening the door to improved forms of \textit{abstractive} summarization---a form of summarization particularly well-suited for multi-party conversation.  In this paper, we provide an overview of the challenges raised by the task of abstractive meeting summarization and of the data sets, models and evaluation metrics that have been used to tackle the problems. 

\end{abstract}

\section{Introduction}

Being a primary and inevitable means of information exchange at the workplace, a vast amount of time and organizational resources are allocated to meetings \citep{mroz2018we}.  \citet{rogelberg2007science} reported that on average, American employees and managers put 6 and 23 hours per week, respectively, into meetings. The rise in videoconferencing linked to the COVID-19 pandemic has only made the situation more severe \citep{kost2020you}: people are having longer and more frequent meetings,  leading to increased fatigue and less time to digest the information exchanged \citep{fauville2021zoom}. 
In this context, developing a system that could reliably identify key information from a meeting or meeting transcript and use it to produce a  condensed and easily digestible summary---as well as, perhaps, a set of meeting minutes that details decisions and action items---is becoming more of a priority than ever before \citep{edmunds2000problem,elciyar2021overloading}. 

Research on automatic summarization dates as far back as  the 1950s with systems aiming to generate abstracts from scientific literature \citep{luhn1958automatic}. Since then, approaches to summarization have developed along two main lines \citep{gambhir2017recent}. \textit{Extractive} summarization, in which a system creates a summary by directly lifting important sentences from source documents and concatenating them without any modification to the original sentences, dominated earlier work due to its simplicity.  The advent of neural encoder-decoder architectures \citep{sutskever2014sequence,vaswani2017attention}, however, has opened the door to \textit{abstractive} summarization 
\citep{see-etal-2017-get,lewis-etal-2020-bart}, which draws upon deep representations of word or sentence meaning to generate well-written, novel sentences that consolidate and concisely paraphrase information that might be distributed among numerous clauses or sentences in the original text or transcript. 

While the majority of research on summarization has been conducted on formal written documents (e.g., news, scientific articles, etc.), in this paper, we highlight work on meeting summarization, a subdomain of automatic summarization that stands to benefit particularly well from advances in abstractive methods.
Meeting summarization poses challenges not encountered by traditional text summarization \citep{kryscinski-etal-2019-neural}. Some are related to limitations on the tools needed for developing models---namely training data, model architectures and evaluation metrics ---but others stem from the very nature of the linguistic interactions involved in meetings or multi-party conversation more generally.  These challenges suggest that producing commercial-level, automatic tools for abstractive meeting summarization cannot be accomplished simply by generalizing or fine-tuning models trained on text or even certain forms of dialogue, but will require developing radically new approaches tailored to the meeting domain \citep{zechner-2002-automatic}.

 We begin in Section \ref{sec:challenges} with an overview of the particular challenges raised by meeting-style interactions and meeting summaries. Sections \ref{sec:data} - \ref{sec:models} introduce the different tools that have been used to address these challenges, focusing on data sets, summary evaluation, and summarization systems, respectively \citep{shang2021spoken}. As we will see, each of these areas introduces complications of its own and we will review the pros and cons of the different approaches. We conclude with a discussion of promising directions for future research.

\section{The challenges of meeting-style speech and summaries} \label{sec:challenges}

\begin{figure*}[ht]
\centering
\includegraphics[scale=0.8]{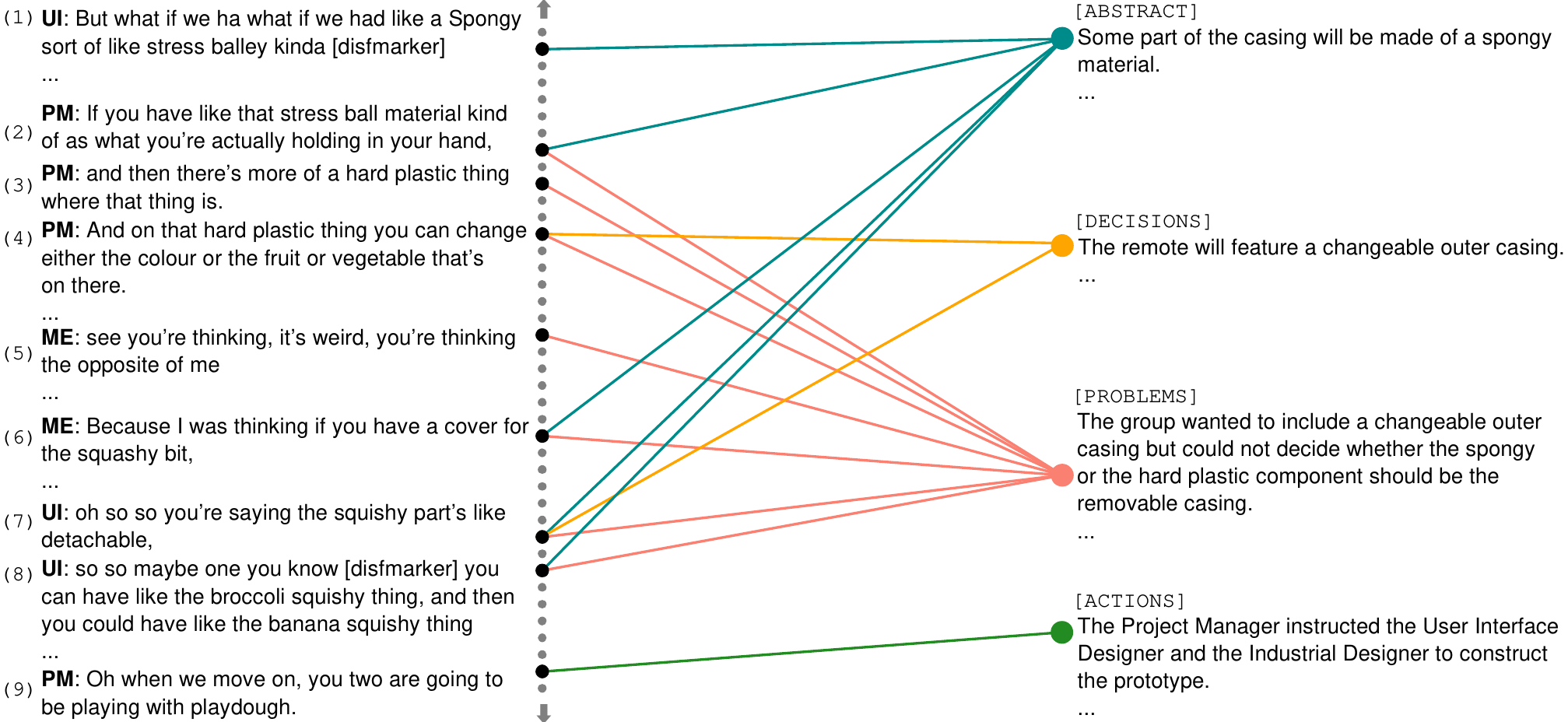}
\caption{Example \citep{shang-etal-2020-energy} of ground truth human annotations from the ES2011c AMI meeting. Successive grey nodes on the left denote utterances recorded in the transcript, the black nodes correspond to a subset of utterances that annotators judged  important (summary-worthy). A single sentence from each section of the abstractive summary is shown on the right. The colored arcs link a given sentence from the abstract summary to a single \textit{abstractive} community; that is, the set of clauses from the extractive summary that is summarized by the sentence from the abstractive summary.}
\label{fig:community}
\end{figure*}

Figure \ref{fig:community} provides a glimpse of the idiosyncratic nature of meeting-style exchanges as well as certain features typical of meeting summaries. The examples are taken from the AMI corpus \citep{mccowan2005ami}, a corpus of interactions in which four participants play the roles of project manager (PM),  marketing expert (ME), user interface designer (UI), and industrial designer (ID) within a fictitious electronics company. The meetings are centered around a design team whose task is to develop a new television remote control from inception to market, through individual work and a series of group meetings (see Section \ref{sec:data} for discussion). Figure \ref{fig:community} shows excerpts of  the human-made extractive (left column) and abstractive (right column) summaries of meeting ES2011c.   The colored lines relate each abstractive sentence to the set of extractive sentences---the \textit{abstractive community}---that annotators judged as supporting it.

\noindent\textbf{The  nature of meeting-style speech.}
Conversations in meetings are unedited, often unprepared  exercises in real-time collaboration, which tends to yield transcripts with low information density and a fair amount of ``noise'' stemming from choppy, repetitive and meandering language. Even though (1) - (9) in Figure \ref{fig:community} are taken from an extractive summary, and thus provide a ``cleaned-up'' selection of utterances from ES2011c, they illustrate some of the disfluent language---e.g., false starts and repetitions (``what if we ha what if we had'') and incomplete utterances (``stress balley kinda [disfmarker]\footnote{``[disfmarker]'' is short for ``disfluency marker'' and covers sounds that were not transcribable.}'')---and long-winded speech (``so so maybe one you know [disfmarker] you can have like the broccoli squishy thing'') common in conversational speech.

Meeting transcripts  also tend to be significantly longer than documents for text summarization, and likewise for their summaries. On average, one AMI transcript contains 4,757 tokens and its summary has 322, while an article from the CNN/DailyMail dataset \citep{hermann2015teaching} has an average of 781 tokens and its summary, 56.

Moreover, in contrast to other types of conversation, such as customer service calls and medical appointments, meetings 
often involve more than two speakers. This increases the number of (potentially conflicting) perspectives to consider (Figure \ref{fig:community}, (5)), and also the variety of speech styles and potential for overlapping speech. Finally, it complicates the task of speaker and addressee identification and thus, the conversion of first and second person pronouns into third person identifiers, as is needed for summarization.

\noindent\textbf{The preference for abstractive summarization.}
Early work on meeting summarization focused on extractive methods \citep{tur2008calo,5410130,riedhammer2008packing,garg2009clusterrank,tixier-etal-2017-combining}.
However, while such approaches might be suitable for traditional documents --- the LEAD-3 baseline, which takes the first three sentences of a document as its summary,  is a good indicator of basic performance \citep{dohare2017text}, for example --- humans tend to prefer abstractive summarization for conversation \citep{murray-etal-2010-generating}. (1) - (9) in Figure \ref{fig:community}, for instance, provide an extractive alternative to the abstractive sentences on the right, but they are arguably less  digestible for a third-party reader than the abstractive alternatives with their concise, third-person perspective on events. 
Abstractive summarization, however, is inherently more difficult than extraction as it involves content synthesis and language generation in addition to the selection of important material.

\noindent\textbf{Heterogeneous meeting formats.} While some meetings, as in Figure \ref{fig:community},  are more focused on problem solving,  other meetings might be more about simply sharing information or brainstorming new ideas.  In other words, different meetings may focus on different kinds of information \citep{nedoluzhko2019towards}, and there is thus no one-size-fits-all approach to summarization. Detailed minutes \citep{fernandez2008identifying,bui-etal-2009-extracting,wang-cardie-2011-summarizing,wang-cardie-2012-focused,wang-cardie-2013-domain} that provide an outline of proposed ideas, supporting arguments, and decisions might be more appropriate for decision-making meetings,  while  topic summaries might be sufficient for information-sharing meetings, and template-based summaries \citep{oya-etal-2014-template}, better suited for well-framed meetings with clear agendas. Such diversity means that there may be a need for a variety of automatic systems to allow for tailoring summaries to particular meeting formats. 

\noindent \textbf{Subjectivity}. Even if we restrict our attention to a single meeting and specify its  summary format, there may be a fair amount of subjectivity in the summary content and style, which complicates the task of summary evaluation (see Section \ref{sec:Evaluation}). Not only are abstractive systems free to reformulate the same content in their own words or style, but also,  what counts as summary-worthy can vary from one annotator to the next. In fact, even one and the same person can produce different results when asked to summarize the same content on two separate occasions  \citep{rath1961formation}.

\section{Meeting data sets}\label{sec:data}

Supervised learning, which remains the standard for meeting summarization, generally requires large amounts of training data and yields language and domain-dependent models. Given the idiosyncratic nature of meetings and their summaries, this means that meeting summarization requires meeting-specific data sets.  In this section, we first discuss what is required to make an appropriate data set and then introduce  extant meeting corpora.

 \subsection{Data  set conception and creation}

A data set appropriate for meeting summarization would need to contain spontaneous, spoken conversations in order to capture the unedited, disfluent speech common to meetings. The conversations would  need to be relatively long, covering multiple points or topics, like a standard meeting. A good portion of the corpus would need to contain more than two speakers, as many meetings do, because conversational dynamics are affected by participants vying for the floor.

Structured meetings in which participants adhere to clear agendas are  easier to summarize, even for humans. At the same time, real-life meetings are not always so well-behaved. Striking a balance between interactions organized enough for summarization to be feasible while remaining fluid enough to be realistic is important for a model meant to generalize to a real-life setting.  This might mean ensuring that participants do not already know each other, in order to minimize small-talk and dependence on  background information inaccessible to summarization models. Or it might mean enforcing very clear goals to encourage participants to stay on topic. 

Which meeting and summary formats should be covered by the corpus will also need to be determined. And a further consideration is that participants should not broach topics that are too private to be  shared outside of their context.

Other concerns arise once the data have been recorded. Automatic speech recognition (ASR) systems can produce transcription errors, for example, that are compounded as we progress through the summarization pipeline, making it risky to skip the laborious task of manual correction. Disfluencies often need to be removed through heavy preprocessing. And in many cases, it is desirable to add additional forms of annotation, such as dialogue act labeling (see Section \ref{subsec:discourse}), which generally require intense effort from trained annotators.

\subsection{Extant data sets}
The significant cost and effort required for producing corpora of meetings and associated summaries or minutes, together with  concerns about the privacy of meeting content, mean that there are very few such data sets available. We know of only three  for English --- AMI \citep{mccowan2005ami}, ICSI \citep{janin2003icsi} and ELITR \citep{elitr-minuting-corpus:2022} --- which together offer around 280 hours of meetings. And only ELITR, which contains roughly 50 hours of meetings in Czech, contains data for a language other than English. We note that all three corpora provide gold transcripts that have been either fully human-produced or human-corrected based on ASR-output to avoid compounded errors from ASR transcripts.

\noindent\textbf{The AMI corpus} \citep{mccowan2005ami}, produced as part of the Augmented Multi-party Interaction project, contains 137 scenario-driven meetings ($\sim$65 hours) ranging from 15 to 45 minutes each. As explained in Section \ref{sec:challenges}, corpus participants play the roles of employees in a fictitious electronics company. The role-playing approach helped to produce a data set with well-structured meetings and a standardized summarization pipeline. It also served to minimize concerns about  privacy.  Note that even though the scenario is artificial, how the participants decide to carry it out is not scripted and so the resulting interactions are spontaneous.  That said, the heavily designed nature of the scenarios and the fact that participants in general did not know each other before participating in the corpus arguably led to overly ``well-behaved'' interactions that risk complicating generalization to real-world contexts. There is also a heavy emphasis on multi-modal interaction (slide presentations, interactions with prototypes) that adds an extra level of complication to the summarization task.

\noindent\textbf{The ICSI corpus} \citep{janin2003icsi} consists of 75 naturally-occurring meetings ($\sim$72 hours) recorded at the International Computer Science Institute. In each (weekly) meeting of around one hour, members from research groups (including undergraduate and graduate students and professors) discuss specialized and technical topics such as natural language processing, neural theories of language, and ICSI corpus related issues. There are six participants on average per meeting. Because the ICSI meetings contain real interactions with people who know each other and already have projects underway, the interactions are inevitably closer to real-life meetings and have clear goals. On the other hand, they contain a lot of technical vocabulary and cover specialized subjects for which participants pull from shared background information that may be inaccessible to summarization algorithms.

\noindent \textbf{The ELITR corpus} \citep{elitr-minuting-corpus:2022}  is a recently released corpus of transcripts for 113 technical project meetings in English, and 53 in Czech, totaling over 160 hours of meeting content. Like ICSI, ELITR contains natural, work-based meetings and so the interactions have similar advantages and drawbacks. We note that unlike ICSI and AMI, the original audio recordings for ELITR are not released and sections of certain meetings are censored due to privacy concerns. 

Both the AMI and ICSI corpora offer multiple levels of annotation including topic segmentation (see Section \ref{subsec:topics}) and dialogue act labeling (Section \ref{subsec:discourse}) in addition to extractive summaries,  abstractive summaries and abstractive communities (see Figure \ref{fig:community}). Abstractive summaries in both corpora follow the same structure with four parts: Abstract (high-level description), Decision, Problems and Actions. Annotators were allowed to make summaries of up to 200 words for each category, but were not obligated to provide summaries for all categories unless they felt it was motivated.
Annotators for the ELITR corpus, by contrast, were not provided with a structure for producing minutes. The results can thus vary widely from one annotator to the next, making the corpus a potentially valuable resource for studying subjectivity in minute and summary creation, though this has not yet been explored.  We publicly release a preprocessed version of the AMI-ICSI\footnote{\tiny \url{https://github.com/guokan-shang/ami-and-icsi-corpora}} and ELITR\footnote{\tiny \url{https://github.com/guokan-shang/elitr-minuting-corpus}} corpora including the aforementioned annotations to foster research on this topic.

\section{Evaluation methods}\label{sec:Evaluation}

As noted in Section \ref{sec:challenges}, the subjectivity of meeting summaries, stemming in part from the preference for abstractive summarization, complicates the evaluation task. Evaluation requires both verifying that the content in a system-based summary follows from the original transcript and measuring the overlap with a gold-standard summary. However, given the expressive freedom encouraged by abstractive approaches, checking for semantic entailment and overlap of summary-worthy content requires deep semantic understanding, not just recognition that the same words are used.

Unfortunately, the ROUGE metric \citep{lin-2004-rouge}, which remains the standard  for both meeting and general text summarization, scores system-produced summaries based purely on surface lexicographic matches with a (usually single) gold summary, making it unideal for assessing abstractive summaries.  If we take the abstractive summary from Figure \ref{fig:community} (``Some part of the casing will be made of a spongy material'') as a gold example, ROUGE would assign a higher score to a system that produces ``Some part of the casing will be made of broccoli'' than one that output ``A portion of the outer layer will be constructed from a sponge-like material,'' even though the latter is a perfect reformulation of the gold summary, while the former says something very different (and false). 

A reasonable way to try to improve over ROUGE would be to take advantage of massive, pretrained, contextual word embeddings and a notion of lexical similarity rather than strict lexical overlap. Some recent efforts pursue this direction \citep{sai2022survey}, 
including BERTScore \citep{zhang2019bertscore} and MoverScore \citep{zhao-etal-2019-moverscore}, which aim to measure the semantic distance between the contextualized mapping of a generated summary and the reference, or BARTScore \citep{yuan2021bartscore}, which calculates the log-likelihood of a summary to have been generated, motivated by the fact that a good summary should have a high probability of being generated from a source text. Building upon these methods, DATScore and FrugalScore \citep{eddine2022datscore,kamal-eddine-etal-2022-frugalscore} incorporate \textit{data augmentation} and \textit{knowledge distillation} techniques to further improve performance and overcome their drawbacks.

Despite their rapid development, recent studies on \textit{meta-evaluation} of these metrics show mixed results. \citet{peyrard-2019-studying} and \citet{bhandari-etal-2020-metrics} compare their performance in the context of document summarization and show that metrics strongly disagree in ranking summaries from any narrow scoring range, e.g., there is no consensus among metrics regarding which summary is better than another in the high scoring range in which modern systems now operate. \citet{bhandari-etal-2020-evaluating} argue that there is no one-size-fits-all metric that correlates better with human judgement than the others, and that can outperform others on all datasets.  

Clearly, evaluation is itself a very challenging task. And we note that none of these metrics even touches on another central challenge for summary evaluation, namely that of factual consistency. When we summarize a meeting or detail decisions and actions items in our own words, it is important to get the facts straight. Otherwise, the resulting summaries are not reliable for an end user. While we do not know of current work that focuses on evaluation of factuality explicitly for the meeting domain, the study of factual consistency in summarization more generally is a budding research area that we discuss in Section \ref{sec:directions}. 

In the absence of a clear winner for summary evaluation metrics, none of the alternatives has yet to be widely adopted. In our comparison of different summarization systems in Section \ref{sec:models}, we therefore stick with ROUGE, which offers the additional advantage of conceptual simplicity and lower computational cost compared to metrics based on contextual embeddings and pretrained language models.

\section{Systems for meeting summarization}\label{sec:models}

\citet{jones1999automatic} describes a general summarization pipeline consisting of three stages: \\
(I) Interpretation: mapping the input text to a \textit{source representation} that adds additional information to the source document that is useful for interpreting its content. \\
(T) Transformation: transforming the source representation to a \textit{summary representation} based on which the final summary will be produced. \\
(G) Generation: generating a summary text from the summary representation.\\
In the course of our literature review, we noted that when extant work on meeting summarization attempts to address one of the challenges described in Section \ref{sec:challenges}, it  can be seen as focusing on one of the stages of Jones' pipeline. We therefore adopt this three-stage schema as the backbone for a taxonomy of summarization systems that we lay out in the remainder of this section.

\begin{table*}[ht!]
\centering
    \ra{1.1}
        \scalebox{0.85}{\begin{tabular}{@{}crrrrcrrr@{}}\toprule
        && \multicolumn{3}{c}{AMI} & \phantom{abc}& \multicolumn{3}{c}{ICSI}\\
        \cmidrule{3-5} \cmidrule{7-9}
        && R-1 & R-2 & R-L && R-1 & R-2 & R-L\\ \midrule
        \multicolumn{1}{c|}{\multirow{4}{1em}{I}}&Sentence-Gated \citep{goo2018abstractive} & 49.29& 19.31& 24.82&& 39.37& 9.57& 17.17 \\
        \multicolumn{1}{c|}{}&DDAMS + DDADA \citep{feng2020dialogue} & 53.15& 22.32& 25.67&& 40.41& 11.02& 19.18\\
        \multicolumn{1}{c|}{}&BART + Discourse \citep{ganesh2019restructuring} & 35.41& 7.24& -&& 31.84& 6.19& -\\
        \multicolumn{1}{c|}{}&TopicSeg + VFOA \citep{li-etal-2019-keep} & 53.29& 13.51& 26.90&& -& -& -\\
        \midrule
        \multicolumn{1}{c|}{\multirow{6}{1em}{T}} & Topic + ILP \citep{banerjee2015generating}& - & 4.80 & - && - & - & - \\
        \multicolumn{1}{c|}{}& Community + MSCG \citep{mehdad-etal-2013-abstractive} & 32.30 & 4.80 & - && - & - & - \\
        \multicolumn{1}{c|}{}&UNS \citep{shang-etal-2018-unsupervised} & 37.86& 7.84& 13.72 && 31.73& 5.14& 14.50 \\
        \multicolumn{1}{c|}{}&PreSeg + POV \citep{park-lee-2022-unsupervised} & 33.66 & 6.85 & 14.17 && 27.80 & 4.56 & 11.77 \\
        \multicolumn{1}{c|}{}&Template-Based \citep{oya-etal-2014-template} & 31.50 & 6.70 & - && - & - & - \\
        \multicolumn{1}{c|}{}&SOAP-Cluster2Sent T5 + HLSTM \citep{krishna-etal-2021-generating} & 50.52& 17.56& 24.89&& -& -& -\\
        \midrule
         \multicolumn{1}{c|}{\multirow{6}{1em}{G}}&SUMM$^N$  \citep{zhang-etal-2022-summn} & 53.44& 20.30& 51.39* && 45.57& 11.49& 43.32*\\
         \multicolumn{1}{c|}{}&Longformer-BART-arg \citep{fabbri-etal-2021-convosumm} & 55.27 & 20.89 & 24.94 && 44.51 & 11.80 & 19.19 \\
        \multicolumn{1}{c|}{}&HMNet  \citep{zhu-etal-2020-hierarchical} & 53.02& 18.57& 24.00&& 46.28& 10.60& 18.54\\
        \multicolumn{1}{c|}{}&HAT-CNNDM  \citep{rohde2021hierarchical} & 52.27& 20.15& 50.57* && 43.98 &10.83 &41.36*\\
        \multicolumn{1}{c|}{}&HAS-RL \citep{zhao2019abstractive} & 48.64 & 17.45 & -&& -& -& -\\
        \multicolumn{1}{c|}{}&DialogLM \citep{zhong2022dialoglm} & 54.49& 20.03& 51.92* && 49.25& 12.31& 46    .80* \\
        \bottomrule
    \end{tabular}}
\caption{Abstractive meeting summarization benchmarks on the AMI and ICSI corpora  \citep[some results are taken from][]{feng2021survey}. The * indicates \textit{summary-level} (with sentence split) ROUGE-L score \citep{lin-2004-rouge}.}
\label{Tab:Benchmark}
\end{table*}

\subsection{Interpretation}
    Consider the following (fabricated) exchange that we might imagine taking place in the AMI meeting from Figure \ref{fig:community}:

\ex.\label{ex:decision}
        \a. PM: So what color should the removable cover be?\label{a}
        \b. ID: I think we should offer a few options. What about raspberry, lime and blueberry and then black for those who don't like color? \label{b}
        \c. UI: Sounds good to me.\label{c}
        \d. ME: Me too.\label{d}
        \e. PM: OK. Let's go with that.\label{e}

    \noindent There is a very clear decision that has been made in \ref{ex:decision}, but what information allows us to recognize this decision so easily? The acknowledgement in \ref{e} explicitly confirms the decision and thus plays a crucial role in inferring that \textit{a} decision has been made, but it does not tell us \textit{what} decision has been made.   
    
    In fact, none of the utterances \ref{a}-\ref{e} alone allow us to infer the decision. We must rather  understand that \ref{b} provides an \textit{answer to the question} asked in \ref{a} and that \ref{e} is \textit{an acknowledgement of the suggestion} in \ref{b} and of \textit{the positive answers} in  \ref{c} and \ref{d}. (If instead of agreeing in \ref{c} and \ref{d}, the UI and ME had presented and defended an alternative proposal, we might have understood \ref{e} as an acknowledgement of \textit{their} suggestion rather than of \ref{b}.)

    Because how an utterance contributes to the larger conversational context is often crucial for understanding  the conversation as a whole, some summarization approaches, which we review in Section \ref{subsec:discourse}, enrich meeting transcripts with explicit representations of those contributions. Other summarization accounts exploit other types of information relevant to discourse interpretation. HMNet \cite{zhu-etal-2020-hierarchical} and DDAMS \cite{feng2020dialogue}, for instance, use information about speakers and turns (``who said what''), drawing on the fact that speaker information can help to  convert (frequent) occurrences of first and second person pronouns into third person pronouns, as is necessary for summarization  \cite{luo-etal-2009-improving}. 
    Further interpretation-focused methods are studied in Section \ref{sec:multi}, in which we take a look at  accounts that have opted to augment transcripts with \textit{non-linguistic} information of multi-modal nature, such as information about the eye-gaze of conversational participants.
    
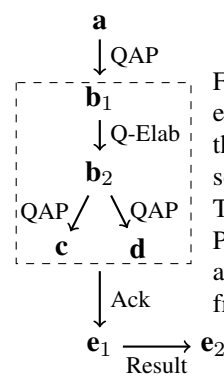
\begin{figure}[ht]
  \centering
\begin{minipage}[c]{0.1\textwidth}
\begin{tikzpicture}
\node (1) at (0,-1.5) {\textbf{a}};
\node (2) at (0,-2.5) {\textbf{b$_1$}};
\node (3) at (0,-3.5) {\textbf{b$_2$}};
\node (4) at (-0.5,-4.5) {\textbf{c}};
\node (5) at (0.5,-4.5) {\textbf{d}};
\node (6) at (0,-5.8) {\textbf{e$_1$}};
\node (7) at (1.5,-5.8) {\textbf{e$_2$}};

\draw[dashed] (-1.1, -2.3) rectangle (1.2,-4.7);

\path[->,thick]
(1) edge node [right] {{\footnotesize QAP}} (2)
(2) edge node [right] {{\footnotesize Q-Elab}} (3)
(3) edge node [left] {{\footnotesize QAP}} (4)
(3) edge node [right] {{\footnotesize QAP}} (5)
(6) edge node [below] {{\footnotesize Result}} (7)
(0,-4.8) edge node [right] {{\footnotesize Ack}} (6)
;

\end{tikzpicture}
\end{minipage}\hfill
\begin{minipage}[c]{0.315\textwidth}  
  \caption{Discourse graph for example \ref{ex:decision}. Node b$_1$ represents the first sentence of \ref{b}, b$_2$, the second, and similarly for e$_1$ and e$_2$. The dashed box indicates that the PM's ``OK'' in \ref{e} acknowledges and accepts the entire exchange from \ref{b} to \ref{d}.\\
  }
  \label{fig:discourse}
\end{minipage}
\end{figure}

    \subsubsection{Discursive information}\label{subsec:discourse}
    While most work on abstractive meeting summarization implicitly assumes that  conversation is merely a linear sequence of utterances without semantic relations between them,  example \ref{ex:decision} underscores the importance of semantic relations for conversational understanding. Drawing on similar insights, certain recent approaches exploit independent theories of \textbf{discourse structure}, such as Rhetorical Structure Theory (RST; \citealp{mann1987rhetorical}) and Segmented Discourse Representation Theory (SDRT; \citealp{asher1993reference,lascarides2008segmented}), to  improve summarization.   Accounts like RST and SDRT maintain that in a coherent conversation, each (roughly) clause-level unit 
    should be semantically related to some other part of the conversation via a \textit{discourse relation} such as Question-Answer Pair (QAP), Acknowledgement (Ack), Explanation, Contrast, etc.~to reflect its contribution to the larger discourse. Each coherent discourse can thus be represented as a weakly-connected graph whose edges are labeled with discourse relations. Figure \ref{fig:discourse} illustrates a possible SDRT graph for example \ref{ex:decision}.

To the best of our knowledge, \citet{feng2020dialogue} is the first work to exploit discourse graphs to generate abstractive meeting summaries. They employ a sequential discourse parser \citep{shi2019deep} trained on the STAC corpus of multi-party chats \citep{asher-etal-2016-discourse} to automatically obtain discourse graphs for the AMI and ICSI meeting corpora. Levi graph transformation \citep{gross2003handbook} is then used to turn graph edges labeled with discourse relation types into  vertices. Their graph-to-sequence model consists of a graph convolutional network encoder \citep{schlichtkrull2018modeling} that takes a meeting discourse graph as input and a PGN decoder \citep{see-etal-2017-get} to generate the final summary. A dialogue-aware data augmentation strategy for constructing pseudo-summaries is introduced to pretrain the model.

  An alternative approach to discourse interpretation that developed largely independently of RST and SDRT is \textbf{dialogue act} classification. Detailing the differences between the  approaches is out of the scope of this paper, but in a nutshell, dialogue acts provide a  shallower notion of discourse structure \citep{jurafsky1997switchboard} in that they do not entail a full  graph structure over a conversation. On the other hand, systems for dialogue act labeling, such as DAMSL \citep{allen1997draft, core1997coding} or DiAML \citep{bunt-etal-2010-towards,bunt-etal-2012-iso}, place more emphasis on interactive acts such as \textit{stalling} to hold the floor, \textit{assessing} other discourse moves, \textit{suggesting} or \textit{informing}. 

  Both the AMI and ICSI corpora provide gold dialogue act labels, and  
    \citet{goo2018abstractive} use the labels from the AMI corpus to develop a sentence-gated mechanism that jointly models the relationships between dialogue acts and topic summaries to improve abstractive meeting summarization. In particular, they show that dialogue acts of the type \textit{inform} are more closely linked to summary-worthy material than acts such as \textit{stall} or \textit{assess}.  Using LSTM, their model consists of three components enhanced with various attention mechanisms: an utterance encoder, a dialogue act labeler, and a summary decoder.

        Dialogue acts have also been used to good effect for summarizing decisions. \citet{fernandez2008identifying} and \citet{bui-etal-2009-extracting} first identify relevant areas of decision-related discussion in meetings and classify them as decision-related dialogue acts including the \textit{issue} under discussion, its \textit{resolution}, or  \textit{agreement} with the proposed resolution. Then, key fragments of the decision related utterances are retained to form an extractive decision summary. Similar ideas can also be found in the literature on detecting and summarizing action-item-specific dialogue acts in meetings \citep{purver2006detecting,purver-etal-2007-detecting}.

    \subsubsection{Multimodality}\label{sec:multi}

    Meetings tend to take place in shared visual contexts, allowing important information to be conveyed through gestures, facial expressions, head poses, eye gaze, and so on. It is thus reasonable to think that harnessing this information could lead to improved performance over a summarization system that draws on speech transcripts alone. Indeed, \citet{li-etal-2019-keep} has shown that the Visual Focus Of Attention (VFOA), which is estimated based on a participant’s head orientation and eye gaze, can help a multi-modal summarization system determine salient utterances, thereby improving the quality of abstractive meeting summaries.  In this work, VFOA is integrated with attention mechanisms in a PGN text decoder and topic segmenter trained in a multi-task learning setting.

    In the context of extractive meeting summarization,  \citet{erol2003multimodal} exploits multimodality to select segments of  meeting video recordings in order to facilitate browsing of the videos for important information.   Along with TF-IDF text analysis of the speech transcript, their system attends to certain visual motion activities (e.g., someone entering the meeting room or standing up to make a presentation) and to acoustic features (e.g., speech volume and degree of interaction) to enhance  detection of summary-worthy events. Similar ideas can  be found in   \citet{xie-liu-2010-using,nihei2016meeting,nihei2018fusing,nihei2019exploring}.

    Of course, enhancing meeting transcripts with information to improve interpretation comes at a cost: most annotations are performed manually and require linguistic expertise to do well, and annotations produced automatically, as in \citet{shi2019deep} or \citet{shang-etal-2020-speaker}, can add uncertainties to the data. It is perhaps thus unsurprising  that the last couple of years have seen a shift away from  Interpretation-focused methods presented in this section  towards Generation methods, as shown in Table \ref{Tab:Benchmark}. Still, certain approaches showed promising results, especially \citet{feng2020dialogue} and \citet{li-etal-2019-keep}, and arguably, Interpretation approaches offer other advantages that should encourage us to explore them further. Discourse graphs, for example, can provide input to graph-based networks, helping to bypass limits on input document length imposed by some Generation approaches (see Section \ref{subsec:generation}). And a rich source representation can arguably make a single transcript more valuable for training, partially offsetting data scarcity described in Section \ref{sec:data}.

\subsection{Transformation}

Similar in spirit to many standard multi-step, text summarization techniques \citep{salton1997automatic} that begin by grouping sentences according to semantic similarity,
this category of work focuses  on transforming  meeting transcripts into intermediate representations that make them easier to summarize. 
Utterances are first grouped according to various criteria, such as whether they share a topic,  contribute to the same field in a template, or respond to the same query.
The motivating idea is that first segmenting a transcript into chunks will help to streamline summary generation, as each chunk will be focused on a different aspect of the meeting. These approaches often add an extractive summarization component as well that is applied either before or after chunking to filter out non summary-worthy utterances.

As explained in section \ref{sec:challenges}, different meetings require different types of summaries;  patient doctor consultations, for example, require SOAP notes to be produced to summarize the consultation point by point \citep{krishna-etal-2021-generating}. Having an initial segmentation creates versatile representations that can be used by summarization systems to output domain specific summaries that  follow templates or topics. Additionally, while the lengthy nature of meetings is one of the biggest challenges of summarization, segmenting a meeting into multiple, topic focused sub-parts creates units small enough to be easily processed by systems that are computationally expensive.

\subsubsection{Topic segments}\label{subsec:topics}
Topic segmentation involves dividing text into topically coherent segments of sentences.
With the help of a topic segmenter, such as LCSeg \citep{galley-etal-2003-discourse}, the work of \citet{banerjee2015generating} first separates a meeting transcript into several topic segments. The dependency parse trees of all of the sentences in a topic segment are then merged into a global dependency graph from which the most informative and well-formed sub-graph (i.e. sentence) is selected. Finally, the selected sentences are  combined  to generate a  summary. 
In the larger dialogue domain, topic segments are also useful for summarizing nurse-to-patient dialogues \citep{liu2019topic} and unstructured daily chats \citep{chen-yang-2020-multi}.

\subsubsection{Abstractive communities}
Introduced in \citet{murray-etal-2012-using}, \textit{abstractive community detection} is a sub-task of abstractive meeting summarization that seeks to identify the clusters of extractive sentences, or abstractive communities, that together support an individual sentence from an abstractive summary (see Figure \ref{fig:community}). 
\citet{murray-etal-2012-using} train a logistic regression classifier with handcrafted features to predict if two utterances belong to the same community, then build an utterance graph whose edges represent the binary predictions of the classifier, and finally apply an overlapping community detection algorithm to the graph.
\citet{shang-etal-2020-energy} approach this task by applying Fuzzy c-Means algorithm \citep{bezdek1984fcm} on the utterance embeddings learned with a siamese or triplet network \citep{chopra2005learning,hoffer2015deep}.

\citet{mehdad-etal-2013-abstractive} is the first work to propose an abstractive meeting summarization system built on communities. 
They first extend the method of \citet{murray-etal-2012-using} by using entailment relations to eliminate less informative utterances from each detected community. Then, the utterances of the same community are represented by a multi-sentence compression graph (MSCG, \citealp{filippova-2010-multi}), an unsupervised NLG component to compress a cluster of related, overlapping sentences. The best compression path is extracted as the abstractive summary sentence of the given utterance community.
\citet{shang-etal-2018-unsupervised} and \citet{park-lee-2022-unsupervised} follow the work of \citet{mehdad-etal-2013-abstractive}, introducing various novel components to improve the original summarization process. Abstractive communities of specific types (e.g., \texttt{decisions}) have also proven useful in summarizing decisions \citep{wang-cardie-2011-summarizing,wang-cardie-2012-focused}.

\subsubsection{Template-related clusters}
In some cases, a meeting summary or the sentences therein are expected to follow a specific template. 
\citet{oya-etal-2014-template} introduces the first template-based abstractive meeting summarization system. Using a clustering algorithm and the MSCG, it creates a set of sentence templates generalized from human-authored abstractive sentences, e.g., ``[speaker] discussed [act] and [content]''. For a given meeting to be summarized, the system first segments the transcription based on topics, then finds the best sentence templates for each segment, and fills the templates to create summaries.
The work of \citet{krishna-etal-2021-generating} leverages template structure at the summary-level, e.g., the four parts of an AMI summary. Using a multi-label classifier, the system first extracts noteworthy utterances, while also predicting the section(s) for which they are relevant, i.e., clustering utterances with respect to the in-template sections. Finally, it generates the final summary in a section-by-section way, using only each section’s predicted noteworthy utterances.
This approach is also tested on clinical summary generation of doctor-patient conversations. 

\subsubsection{Query-related clusters} \label{subsubsec:query}

To deal with the problem of heterogeneous meeting formats as well as the subjectivity inherent in summary-production,  \citet{mehdad-etal-2014-abstractive} proposed a query-based meeting summarization system that allows for the creation of different summaries for different user interests expressed via queries. 
Given a phrasal query, the system first extracts the utterances that are most relevant to the keywords of the query and of the main meeting content. Then, these utterances are compressed into summary sentences with an MSCG. Since there were  no human-written query-based summaries for AMI at the time, this system was evaluated manually.
The QMSum dataset \citep{zhong-etal-2021-qmsum} provides multiple query-summary pairs for each meeting in the AMI and ICSI corpora, covering both general and specific points people might be interest in.
They also propose a system with two distinct modules: a locator based on pointer network \citep{vinyals2015pointer} and a neural seq2seq summarizer. 
The former is used to locate the query-related utterances in the meeting transcripts, and the latter is meant to summarize selected utterances into summaries.

While versatile in the tasks they could potentially be used to solve, Transformation-focused systems tend to perform worse than Interpretation and Generation-focused systems (cf.~Table \ref{Tab:Benchmark}). This can be attributed to the relative age of those methods, as well as their conceptual simplicity that focus on extractive and unsupervised summarization \citep{shang-etal-2018-unsupervised, krishna-etal-2021-generating}. The spontaneous nature of meetings can lead to meandering structures and information dispersed over a variety of (not necessarily sequential) turns: participants have side conversations, they forget things and come back, they get interrupted, and so on. All of this makes meeting transcripts harder to accurately segment than traditional documents \citep{xing2021improving}. Finding ways to pre-process a text with an eye to creating high-quality topic-based segmentations to feed to Interpretation or Generation-focused systems could be a way to improve the state of the art.

\subsection{Generation} \label{subsec:generation}

Recently, massive pretrained language models based on transformers  \citep{vaswani2017attention}, such as BART \citep{lewis-etal-2020-bart} and T5 \cite{10.5555/3455716.3455856}, and many others \citep{keskar2019ctrl,NEURIPS2020_1457c0d6,kamal-eddine-etal-2021-barthez}, have brought significant improvement to a variety of NLG tasks. 
Despite their success in dealing with documents, however, these models are inadequate for meeting summarization, if applied off-the-shelf. 
As explained in Section \ref{sec:challenges},  meeting-style speech has an idiosyncratic nature unlike that of the text data on which pretrained language models are trained. In addition, pretrained language models impose harsh limits on input length that fall far short of the length of an average meeting. 
       
   \subsubsection{Long input processing}

    A straightforward solution to the length problem is to segment a long document into smaller segments to be processed. \citet{koay-etal-2021-sliding} separate ICSI meetings with a sliding window, and then apply BART on each segment to produce smaller summaries, which are concatenated into an overall extractive summary.
    \citet{zhang-etal-2022-summn} propose SUMM$^N$, a multi-stage split-then-summarize framework. Within each stage, it first splits the source input into sufficiently short segments. Coarse summaries are then generated for each segment and then concatenated as the input to the next stage. This process is conducted repetitively until a final, fine-grained  abstractive meeting summary is produced.

    While such segmenting approaches address the length problem, they can lose important cross-partition information \citep{beltagy2020longformer}, a risk that has led researchers to seek more sophisticated solutions to the length problem.
    
    \noindent\textbf{Long-sequence transformers.} Multiple variants of adapting transformer-based approaches to address the lengthy input problem exist in the literature \citep{dai-etal-2019-transformer,beltagy2020longformer,martins-etal-2022-former}.  Longformer \citep{beltagy2020longformer}, for example, introduces a multi-layer self-attention operation that scales linearly with sequence length, enabling it to process long meeting transcriptions. Although these models are not initially proposed for abstractive meeting summarization, recent benchmarks show their promise \citep{fabbri-etal-2021-convosumm} in comparison with certain dedicated systems.
   
    \noindent\textbf{Hierarchical transformers.} Some systems leverage transformers in a hierarchical manner, breaking down a long meeting transcript to multiple relatively shorter sequences of different levels, mirroring the underlying hierarchical structure of text \citep{yang-etal-2016-hierarchical}, i.e., words combine into an utterance, and utterances form a transcription. 
    The HMNet model, proposed by \citet{zhu-etal-2020-hierarchical}, follows a two-level structure. First, each utterance in the meeting is separately encoded by the same word-level transformer encoder, resulting in a sequence of utterance vectors. That sequence is then processed by the turn-level encoder. The transformer decoder makes use of both levels of representation via cross-attention layers.
    \citet{rohde2021hierarchical} propose Hierachical Attention Transformer (HAT). Utterances are first prepended with a special BOS token. Then, after obtaining token-level embeddings with a standard transformer encoder, the BOS token embeddings are fed into an extra layer, yielding sentence-level representations. Finally, the decoder leverages the outputs at both levels to produce a final summary.
    Similarly, the hierarchical encoder of \citep{zhao2019abstractive} consists of three levels, sequentially encoding word, utterance and topic segment embeddings.

    \subsubsection{Domain adaptation}

    \citet{gururangan-etal-2020-dont} have shown that instead of directly using an off-the-shelf language model pretrained on a massive, heterogeneous, and broad-coverage corpus, it is helpful in performance gains to conduct domain-adaptive and task-adaptive pretraining.
    
    \citet{zhong2022dialoglm} present the DialogLM model, along with a dialogue-dedicated, window-based denoising pretraining approach.  Windows of consecutive utterances are first selected from the conversation, which is then disrupted with arbitrary dialogue-related noises, e.g., speaker mask, turn splitting, turn merging, text infilling and turn permutation. In the end, the model is trained with the objective of reconstructing the original window based on the perturbed one and the remaining conversation.
    This approach allows a model to effectively learn dynamic dialogue structures within and surrounding the window.
    \citet{zou-etal-2021-low} propose a multi-source pretraining paradigm for low-resource dialogue summarization. They first pretrain the encoder and the decoder separately on the in-domain data, to model the dialogue and summary language respectively, and then pretrain the complete encoder-decoder on the out-of-domain abstractive summarization data using adversarial critics.
   
    Generation-focused systems are based on pretrained language models, which excel at abstractive summarization, and use their extensive pre-training to offset the lack of task specific data. Domain adaptation also helps these models to understand the idiosyncratic style of multi-party speech, albeit to a lesser extent than Interpretation-focused systems.  Generation-focused systems, however, lack the flexibility that Interpretation-focused systems have in their output, and when we consider how competitive scores are for Interpretation-focused systems despite their having considerably less training data than Generation-based systems, it is arguable that the latter lack the powerful understanding in the initial representation exhibited by the former. In the following section, we will talk about how some of these shortcomings can be addressed in future work.

\section{Future directions}\label{sec:directions}

We conclude by laying out some directions that we think are worth  investigation in future work.

\noindent\textbf{Multi-task learning.}
    Given the variety of annotations that have led to improvements in abstractive meeting summarization, including dialogue act classification, discourse parsing, community detection, multimodal information and so on, we might expect to see more systems combining these tasks with meeting summarization. 
    Studies on multi-task learning have shown that solving different tasks in a simultaneous fashion often improves learning efficiency and performance, compared to  models trained separately \citep{caruana1997multitask}, potentially reducing the amount of data needed to solve an individual task. Future work could follow up on previous attempts in this direction \citep{li-etal-2019-keep,feng-etal-2021-language,lee2021speaks}.

\noindent\textbf{Factual consistency.}
Information inconsistency is a common problem of summarization systems \citep{kryscinski-etal-2019-neural,kryscinski-etal-2020-evaluating}, 
and it is reported that nearly 30\% of summaries generated by neural seq2seq models suffer from fact fabrication \citep{cao2018faithful}.
This problem is not only present in the generated summaries, but also in the training data \citep{guo-etal-2022-questioning}.
For the dialogue domain, \citet{tang-etal-2022-confit} show that most of the factual errors are related to dialogue flow modeling, informal interactions between speakers, and complex coreference resolution. 
This suggests future research to focus on dealing with hallucinated content in generated dialogue summaries.

    Moreover, specific attention could be addressed to the development of an evaluation metric that estimates factual consistency between a summary and its source, specifically designed for the dialogue domain given its special characteristics \citep{zechner-waibel-2000-minimizing}. Although many metrics have been put in place \citep{huang2021factual}, they are inadequate for meeting summarization, as they usually work at the sentence level, which is acceptable for shorter documents but hardly for longer meeting transcripts. 

\noindent\textbf{Weak to no supervision.}
    Current work on meeting summarization focuses mainly on supervised approaches, which makes trained models inevitably language-dependent, and even corpus-dependent. One way to get around the data scarcity problem might be to move to meeting summarization techniques with weaker supervision, such as in a purely unsupervised fashion \citep{shang-etal-2018-unsupervised,park-lee-2022-unsupervised}, or even in a zero-shot way \citep{ganesh2019restructuring}. 
    Future approaches could leverage auto-encoders and utterance wide attention to identify the importance of different parts of a meeting. Weak supervision and semi supervised approaches have yet to be explored in detail.

\noindent\textbf{Spoken language models.}
    Over the past few years, pretrained language models have evolved from emerging to mainstream NLP technology. As mentioned in Subsection \ref{subsec:generation}, models pretrained on free-form text cannot be directly transferred to treat dialogue transcriptions.
    Future research could follow the work of, notably, DialoGPT \citep{zhang-etal-2020-dialogpt} and DialogLM \cite{zhong2022dialoglm}, to develop better language models dedicated to spoken language,
    which will potentially offer large gains in task performance for abstractive meeting summarization.

\noindent\textbf{Commonsense incorporation.}
    Some facts are naturally considered as commonsense knowledge for human beings, but are far from obvious for machines, e.g., ``a piano is played by pressing keys''.
    The integration of such \textit{a priori} knowledge, usually represented by graphs, into language models would improve their interpretation and reasoning potential \citep{ilievski2021cskg}, thereby further  combining  the strength of Interpretation and Generation-focused systems.
    Commonsense knowledge has already been applied to general dialogue summarization with success \cite{zhou2018commonsense,xiachong-etal-2021-incorporating}. For meetings, we believe that its incorporation, much like explicitly flagging jargon terms \citep{koay-etal-2020-domain}, would provide further information that is likely not included in the context. Indeed, considering that meetings take place in professional environments, some background concepts are taken as given by the participants, and integrating this knowledge into  models would undoubtedly facilitate dialogue understanding and summary generation.

\noindent\textbf{Prompting paradigm.}
The very recent success of zero- and few-shot learning with models like GPT-3 \citep{NEURIPS2020_1457c0d6}, Gopher \citep{rae2021scaling}, and PaLM \citep{Chowdhery2022PaLMSL} provides a novel paradigm in NLP research.
In contrast to the traditional approach of fine-tuning pre-trained language models for tasks using task-specific data sets, this approach yields high performance on a variety of NLP tasks without  updating parameters. All you have to do is to give the models task-specific instructions (namely prompts) represented in natural language, possibly in conjunction with a few demonstrative examples in the context.
For example, a prompt for obtaining the summary of an \textit{article} given a budget constraint of $N$ sentences can be formulated as:\\
   \indent ``Article: \{\{article\}\}''\\
   \indent ``Summarize the above article in N sentences.''\\
Recent work shows that these models can be further enhanced by instruction tuning (e.g., T0 \citep{DBLP:journals/corr/abs-2110-08207}, FLAN \citep{wei2021finetuned}, and BLOOMZ \citep{scao2022bloom}) and learning from human feedback (e.g., InstructGPT/ChatGPT \citep{ouyang2022training}), in order to be better aligned with inference time usage.

Experiments on the traditional summarization datasets (e.g., CNN/DailyMail) show that these models can produce summaries nearly as good as reference summaries \citep{stiennon2020learning}, and humans overwhelmingly prefer the summaries produced under the novel prompting paradigm over the classical pretraining-finetuning one \citep{goyal2022news}. To the best of our knowledge, there has been no attempt to apply it to meeting summarization, but this is an exciting avenue for future research.

Moreover, \citet{goyal2022news} pointed out that this fundamental paradigm shift brings new and serious challenges for evaluation: both existing reference-based and reference-free automatic metrics cannot reliably evaluate zero-shot summaries. 
Because prompt-based models are not restrictively trained to emulate gold-standard summaries and their style,  reference-based metrics that compute  lexical or semantic similarity (e.g., ROUGE and BERTScore) give false low scores when comparing generated summaries against available gold summaries. Reference-free metrics (e.g., BLANC \citep{vasilyev-etal-2020-fill} and FactCC \citep{kryscinski-etal-2020-evaluating}) that only rely on the input document also fail, as reference-free metrics roughly score summaries inversely related to abstractiveness, while prompt-based models often generate more abstractive summaries.
Therefore, new metrics adapting to this paradigm shift that can evaluate zero-shot summaries are urgently needed.

In the end, as mentioned in Section \ref{sec:challenges}, there is arguably no single ``correct'' summary for a given input meeting. Different users may choose to focus on different information, choosing different keywords and mentioning different facts. With prompt-based models, the task can be comfortably formulated by creating optimized prompts to meet different needs. Therefore, a study about such system, as well as a comparison with existing query-based meeting summarization systems (see Section \ref{subsubsec:query}) for the same purpose could be an interesting research topic.

\section{Conclusion}

    In this work, we presented a comprehensive overview of the state-of-the-art in abstractive meeting summarization. We discussed the different challenges that research has faced, and we proposed a taxonomy that followed the three-step summarization pipeline, Interpretation-Transformation-Generation, presented by \citet{jones1999automatic}. We further compared the results obtained by  previous work and finally laid out  promising directions for future research.

\section*{Acknowledgments}

    This work was supported by the SUMM-RE project (ANR-20-CE23-0017). We thank the anonymous reviewers for their feedback.

\bibliography{googlescholar,aclanthology}
\bibliographystyle{acl_natbib}

\end{document}